\documentclass{article}

\usepackage{microtype}
\usepackage{subfigure}
\usepackage{natbib}
\usepackage{graphicx}
\usepackage[accepted]{arxiv_icml} 
\usepackage{booktabs}
\usepackage{amsmath}
\usepackage{amsfonts}

\usepackage{hyperref}

\DeclareMathOperator*{\argmin}{argmin} 

\icmltitlerunning{DIP for Univariate Time Series Inverse Problems}

\begin{document}

\twocolumn[
\icmltitle{One-dimensional Deep Image Prior \\ for Time Series Inverse Problems}

\begin{icmlauthorlist}
\icmlauthor{Sriram Ravula}{UT}
\icmlauthor{Alexandros G. Dimakis}{UT}
\end{icmlauthorlist}

\icmlaffiliation{UT}{The University of Texas at Austin, Department of Electrical and Computer Engineering}

\icmlcorrespondingauthor{Sriram Ravula}{Sriram.Ravula@utexas.edu}
\icmlcorrespondingauthor{Alexandros G. Dimakis}{dimakis@austin.utexas.edu}

\icmlkeywords{Compressed Sensing, Deep Image Prior, Time Series}

\vskip 0.3in
] 

\printAffiliationsAndNotice{}

\begin{abstract}

We extend the Deep Image Prior (DIP) framework to one-dimensional signals. DIP is using a randomly initialized convolutional neural network (CNN) to solve linear inverse problems by optimizing over weights to fit the observed measurements. 
Our main finding is that properly tuned one-dimensional convolutional architectures provide an excellent Deep Image Prior for various types of temporal signals including audio, biological signals, and sensor measurements. We show that our network can be used in a variety of recovery tasks including missing value imputation, blind denoising, and compressed sensing from random Gaussian projections. The key challenge is how to avoid overfitting by carefully tuning early stopping, total variation, and weight decay regularization. 
Our method requires up to 4 times fewer measurements than Lasso and outperforms NLM-VAMP for random Gaussian measurements on audio signals, has similar imputation performance to a Kalman state-space model on a variety of data, and outperforms wavelet filtering in removing additive noise from air-quality sensor readings.    
\end{abstract}

\section{Introduction}

We are interested in reconstructing an unknown signal denoted by $x \in \mathbb{R}^n$, after observing linear projections on its entries
\begin{equation}
    y = Ax + \eta.
\end{equation}
Here, the vector $y \in \mathbb{R}^{m}$ corresponds to the observed measurements, 
$A \in \mathbb{R}^{m\times{n}}$ is the measurement matrix and $\eta \in \mathbb{R}^{m}$ is the additive measurement noise. For many applications, the number of measurements $m$ is smaller than the unknown signal dimension $n$,
which results in an ill-posed problem: simply put, there are many possible signals that explain the measurements. 

For these high-dimensional problems, the usual solution is to leverage additional knowledge on the structure of the unknown signal. 
The most common such assumption is sparsity, which leads to regularization with the $l_1$ norm and the widely used Lasso, see e.g.~\cite{tibshirani2011regression, Candes2006,donoho2006compressed} and the significant volume of more recent work.
Beyond sparsity, several complex models such as deep generative models, mesh projections and model-based compressed sensing have been effective in recovering signals from underdetermined systems of measurement, e.g. \cite{Bora2017, Gupta2019, Chang2017, Baraniuk2010, Mousavi2019}. 

Deep learning reconstruction methods are very powerful but require training on large datasets of ground-truth images.  
A groundbreaking recent study by Ulyanov et al. demonstrated the ability of Deep Image Prior (DIP), an \textit{untrained} convolutional neural network (CNN), to perform image inverse tasks like denoising and super-resolution (\citeyear{Ulyanov2018}). This study showed that a randomly initialized convolutional neural network can be optimized \textit{over its weights} to produce an image closely resembling a target image. Since the procedure uses information from only a single sample, traditional network training is not required.
Very recently, deep image prior was extended to general inverse problems for imaging using two dimensional convolutional neural network architectures~\cite{VanVeen2018}. 

Beyond imaging, one-dimensional time series data recovery is another field that has seen advances due to deep learning. Recently, neural models have been proposed in the area of imputing missing multivariate time series data \cite{Luo2018, Che2018, Cao2018}. These models rely on learning correlations between variables as well as time relations between observations to produce informed predictions. Univariate time series
present a more challenging problem: since we only 
have information about a single signal in time, only temporal patterns can be exploited to recover the original missing information~\cite{Moritz2015}. 
Therefore, algorithms must exploit knowledge about the structure of the natural time signals being sensed.

\begin{figure}[ht]
\vskip 0.2in
\begin{center}
\centerline{\includegraphics[width=\columnwidth]{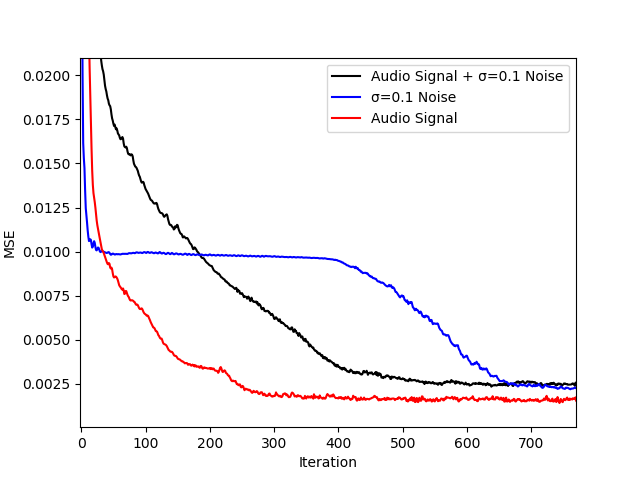}}
\caption{
Optimizing the weights of our network to make it fit Gaussian noise, an audio signal, and their sum, as a function of training iterations. 
Observe that the CNN weight optimization can fit on reconstructing the natural signal much faster compared to reconstructing the noise. 
This is the first indication that the one-dimensional convolutional architecture is a good prior for natural univariate time series and also shows the importance of early stopping. }
\label{noise_fitting}
\end{center}
\vskip -0.2in
\end{figure}

In this paper we develop a deep image prior methodology to solve inverse problems for univariate time series. Our central finding is that one-dimensional convolutional architectures are excellent prior models for various types of natural signals including speech, biological audio signals, air sensor time series, and artificial signals with time-varying spectral content. Figure~\ref{noise_fitting} demonstrates how DIP has high resistance to noise but converges quickly when reconstructing a natural signal.
This is quite surprising and the exploited structure is not simply low-pass frequency 
structure: if this was the case, Lasso in DCT would outperform our method. 
Instead, the one dimensional convolutional architecture enforces time-invariance and manages to capture natural signals of different types quite well.

As with prior works, the central challenge in deep image prior methods is correctly regularizing:
unless correctly constrained, optimizing over the weights of a convolutional neural network to match observations from a single signal will fail to yield realistic reconstructions. 
Our innovations include designing a one-dimensional convolutional architecture and optimizing early stopping as well as utilizing total variation and weight decay regularization. 

We demonstrate that our CNN architecture can be applied to recover signals 
under four different measurement processes: Gaussian random projections, random discrete cosine transform (DCT) coefficients, missing observations, and additive noise. 
Our method requires up to $4\times$ fewer measurements compared to Lasso and D-VAMP with the non-local means (NLM) filter and Wiener filter to achieve similar test loss with random Gaussian projections on audio data, when the number of measurements is small. 
For a large number of Gaussian random projections, our method performs worse compared to the state of the art NLM-VAMP \cite{Metzler2016, Rangan2017}. 
For DCT projections the proposed algorithm nearly matches or outpeforms all baselines depending on the signal type. 
Finally, for imputation tasks, our method performs similarly to a Kalman state-space model and outperforms wavelet filtering for blind denoising on air quality data.  

The remainder of this paper is organized as follows. We first briefly discuss the related prior work in deep learning for inverse problems and time series recovery. Then, we describe our methodology, the datasets used and the reconstruction experiments we performed. Finally, we discuss our results and compare the performance of the previous state of the art methods.

\section{Related Work}
Most recent studies concerning time series recovery with deep learning methods deal with imputing missing values in \textit{multivariate} time series. Luo et al. use a generative adversarial network (GAN) to fill in missing air quality and health record observations by optimizing first over the network weights, then over the latent input space to produce an output which matches the known observations as closely as possible (\citeyear{Luo2018}). Che et al. opt to use a recurrent neural network (RNN) with custom gated recurrent units which learn time decays between missing observations to fill observations (\citeyear{Che2018}). Cao et al. similarly demonstrate an RNN's ability to impute missing data by treating missing values as variables of a bidirectional RNN graph and beating baseline methods' performance in classification tasks with the imputed data (\citeyear{Cao2018}). Our work 
is different from these previous studies since we focus on the less-studied area of \textit{univariate} time series. In addition, we use a convolutional network architecture 
as opposed to a recurrent one, to capture time relations.
This limits the applicability of our method to a fixed time-window, or blocks of time for significantly longer time-signals. However, we were able to design architectures for various output sizes without significant problems.

\begin{figure}[t]
\vskip 0.2in
\begin{center}
\centerline{\includegraphics[width=\columnwidth]{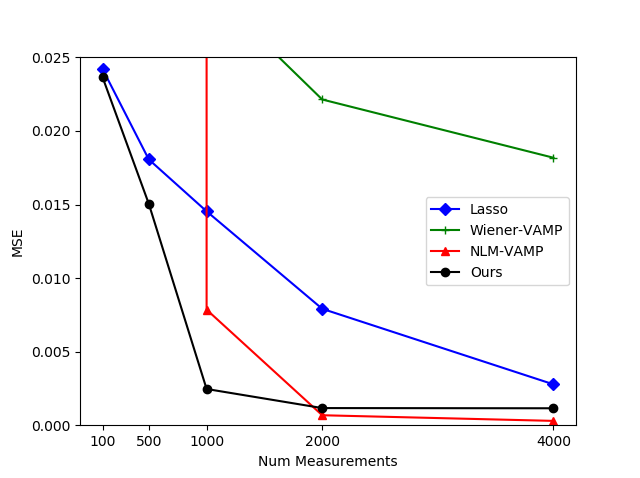}}
\caption{Test loss for recovery from random Gaussian  projections on an audio signal of a drum beat with $n=16,384$ and varying numbers of measurement by our method, Lasso in the DCT basis, NLM-VAMP, and Wiener-VAMP \cite{Metzler2016, Rangan2017}. Our method requires up to $4\times$ fewer measurements than Lasso and outperforms NLM-VAMP for measurements levels below $m=2000$.}
\label{beat_cs}
\end{center}
\vskip -0.2in
\end{figure}

\begin{figure}[ht]
\vskip 0.2in
\begin{center}
\centerline{\includegraphics[width=\columnwidth]{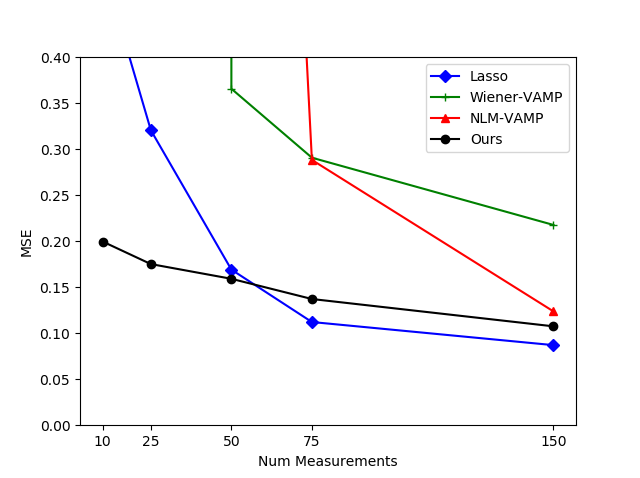}}
\caption{Test loss for recovery from random Gaussian  projections on hourly sensor readings of CO in the air \cite{DeVito2008} with $n=1024$ and varying numbers of measurement by our method, Lasso in the DCT basis, NLM-VAMP, and Wiener-VAMP \cite{Metzler2016, Rangan2017}. Our method produces a more accurate reconstruction than all other methods for levels of measurement below $m=50$, and performs second best to Lasso for greater values. For $m > 200$, NLM-VAMP outperforms all other methods.}
\label{co_cs}
\end{center}
\vskip -0.2in
\end{figure}

Imputation is the most studied research area for univariate time series. Typical methods for recovering missing data include interpolating between known observations, mean-filling, and repeating the last known observation. Moritz et al. describe the 3 main components of univariate signals used in imputation methods as the long-term trend, repeated seasonal shifts, and irregular fluctuations and demonstrate that methods intended for general time series tasks perform worse than methods made specifically to recover one-dimensional series  (\citeyear{Moritz2015}). Phan et al. design such a univariate-specific method to fill in large gaps of successive, non-randomly-missing observations (\citeyear{Phan2017}). Recovering a signal from compressive (i.e. $m<n$) random Gaussian measurements or DCT coefficients are two modes of recovery discussed in our work which are rarely studied for univariate time series signals. In addition, we differ from previous literature by employing a deep generative model to perform  signal recovery.

Much recent work has been published on recovering images from compressively sampled representations. Metzler et. al show that an approximate message-passing (AMP) algorithm which employs a denoiser can be used to recover images from compressed samples, provided that an appropriately effective denoiser is used (\citeyear{Metzler2016}). This approach, called D-AMP, can also be unrolled into a learned version which uses a neural network architecture to learn the model parameters \cite{Metzler2017}. Our model does not depend on the use of an external black box algorithm such as a denoiser and also has the advantage of not requiring training data besides the linear measurements from the signal we wish to recover.

The success of deep learning-based generative networks preceded study into network architecture and regularization.
The result of Deep Image Prior has inspired an under-parameterized deep convolutional generator by Heckel and Hand, who demonstrated that proper network architecture is the crux of optimal signal recovery with DIP (\citeyear{Heckel2018}). This property that the network weights themselves impose a prior on the signal can be called ``regularization by architecture'' \cite{Dittmer2018}. Learned methods of regularization on top of architecture have been proposed which improve performance of compressed sensing with DIP \cite{VanVeen2018}.

The main way that our work diverges from past studies on deep generative methods is that we extend a CNN architecture to one-dimensional signals and implement multiple forms of regularization to properly reconstruct signals. 
In contrast to the work by~\cite{Bora2017} and the follow-up work, we do not need a pre-trained generative model to solve inverse problems. Our method can be applied to a single time series signal to perform imputation, denoising or reconstruction from projections.

\begin{table*}[t]
\caption{Test loss for reconstructing a signal from $m=100,500,1000,2000,$ and $4000$  DCT coefficients of three different audio signals with $n = 16,384$. We compare our method to Lasso in the DCT basis, NLM-VAMP, and Wiener-VAMP \cite{Metzler2016, Rangan2017}. The best MSE value for each test is bolded. DNC indicates that the algorithm did not converge.}
\label{DCT-table}
\vskip 0.2in
\begin{center}
\begin{small}
\begin{sc}
\begin{tabular}{@{}llllll@{}}
\toprule
Method      & \multicolumn{5}{c}{Beat}                                                                                                                                                                         \\\cmidrule(lr){2-6}
            & \multicolumn{1}{c}{$m =100$}         & \multicolumn{1}{c}{$m =500$}         & \multicolumn{1}{c}{$m =1000$}        & \multicolumn{1}{c}{$m =2000$}        & \multicolumn{1}{c}{$m =4000$}        \\\midrule
Ours         & \textbf{0.0231} & \textbf{0.0224} & \textbf{0.0220} & 0.0183 & \textbf{0.0154} \\
Lasso       &0.0238          & 0.0232          &0.0231          & \textbf{0.0177}          & 0.0176          \\
NLM-VAMP    & DNC                                & DNC                                  & DNC                                  & DNC                                  & DNC                                   \\
Wiener-VAMP & 0.0558          & 0.0449          & 0.0272          & 0.0202          & 0.0176          \\\midrule
            & \multicolumn{5}{c}{Whale}                                                                                                                                                                        \\\cmidrule(lr){2-6}
            & $m =100$                             & $m =500$                             & $m =1000$                            & $m =2000$                            & $m =4000$                            \\\midrule
Ours         & 0.0754                              & \textbf{0.0713}                     & \textbf{0.0708}                     & 0.0308                              & 0.0241                              \\
Lasso       & \textbf{0.0746}                     & 0.0743                              & 0.0734                              & \textbf{0.0267}                     & \textbf{0.0227}                     \\
NLM-VAMP    & DNC                                  & DNC                                  & DNC                                  & DNC                                  & DNC                                  \\
Wiener-VAMP & DNC                                  & 0.0973                              & 0.0820                              & 0.0464                              & 0.0242                              \\\midrule
            & \multicolumn{5}{c}{Speech}                                                                                                                                                                       \\\cmidrule(lr){2-6}
            & $m =100$                             & $m =500$                             & $m =1000$                            & $m =2000$                            & $m =4000$                            \\\midrule
Ours         & 0.0811                              & 0.0810                              & 0.0798                              & 0.0744                              & 0.0618                              \\
Lasso       & \textbf{0.0778}                     & \textbf{0.0766}                     & \textbf{0.0746}                     & \textbf{0.0690}                     & \textbf{0.0594}                     \\
NLM-VAMP    & DNC                                  & DNC                                  & DNC                                  & DNC                                  & DNC                                  \\
Wiener-VAMP & DNC                                  & 0.1215                              & 0.0928                              & 0.0727                              & 0.0597\\                             
\bottomrule
\end{tabular}
\end{sc}
\end{small}
\end{center}
\vskip -0.1in
\end{table*}

\section{Background}
We wish to solve the inverse problem of recovering a true signal $x \in \mathbb{R}^{n}$, which has been transformed by a linear measurement process given by matrix $A \in \mathbb{R}^{m\times{n}}$ and summed with noise $\eta \in \mathbb{R}^{m}$. Thus, given the measurements $y = Ax + \eta$ we aim to produce a signal $x^*$ that is as similar to $x$ as possible. 

Ulyanov et al. demonstrated that a deep generative CNN architecture is well-adapted to model natural images, but not noise (\citeyear{Ulyanov2018}). It was shown that even with randomly initialized weights, a CNN could reproduce a clear, natural-looking image without needing to be trained on a large dataset beforehand. This result is remarkable, as DIP only requires optimization over network weights while keeping the latent input fixed. The DIP optimization problem can be formulated as 

\begin{equation}
    w^{*} = \argmin_w \|y - G(z,w)\|^{2},
\end{equation}

where $y \in \mathbb{R}^{n}$ is the observed image and $G(z, w) = \hat{x}$ is the CNN output given latent vector $z$ and network weights $w$. For this case the measurement matrix $A$ is the $n\times{n}$ identity and there is no additive noise. Throughout DIP optimization the latent vector $z$ is kept to fixed to some random initial value as we optimize over the weights $w$.  

\section{Methods \& Experimental Setup}
\subsection{Convolutional Neural Network Setup}
\label{CNNsection}

Our approach is to optimize the output $G(z,w)$ of a CNN so that measurements taken from this output match the observed measurements of $x$, the original univariate time series we wish to recover, as closely as possible. We regularize the network output by minimizing total variation loss, which has shown improvements over simple mean-squared error (MSE) loss optimization for DIP signal recovery tasks \cite{Liu2018}. The optimization task for our method for a given input $z$ therefore is

\begin{equation}
\label{Tv_loss}
w^{*} = \argmin_w \big\{ \|y-AG(z,w)\|^{2} + \lambda \rho_{TV}(G(z,w)) \big\},
\end{equation}

for total variation loss $\rho_{TV}$ given by:

\begin{equation}
\label{TV}
    \rho_{TV}(x) = \sum_{i=2}^{n} |x{[i]} - x{[i-1]|},
\end{equation}

where $\lambda \in \mathbb{R}$ is a tuning parameter to control the amount of regularization by total variation. In other words, our network produces an $n$-dimensional output $G(z,w)$, we perform linear measurement process given by $A \in \mathbb{R}^{m\times{n}}$ on this output to simulate $m$-dimensional measurements, then we optimize the loss given by Equation~\ref{Tv_loss} between our simulated measurements $AG(z,w)$ and the observed measurements $y=Ax$. Though this is a nonconvex problem due to the complexity of $G$, we can still solve this problem using gradient descent. Our aim is that our final network output $x^* = G(z,w^*)$ matches the original signal x as closely as possible. 

We use a one-dimensional convolutional neural network architecture with 64 convolutional filters per layer throughout. We use the PyTorch implementation of the RMSProp optimizer with a learning rate of $10^{-4}$, momentum of 0.9, and weight decay of 1 to optimize our network \cite{paszke2017}. We chose a $\lambda$ of 0.1 for our total variation loss. We chose these parameter values by a grid search over a fixed set of possible values. 

\begin{figure}[t]
\vskip 0.2in
\begin{center}
\centerline{\includegraphics[width=\columnwidth]{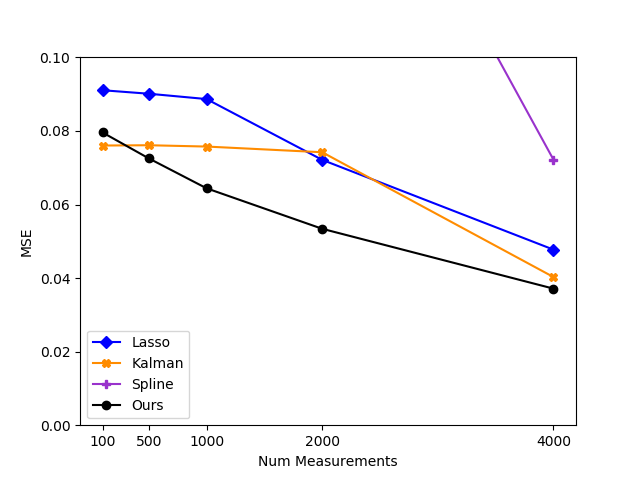}}
\caption{Test loss for imputation on an audio signal of speech with $n=16,384$ and varying numbers of measurement by our method, Lasso in the DCT basis, Kalman state-space imputation, and spline interpolation \cite{Moritz2017}. Our method outperforms Kalman interpolation, requiring up to $4\times$ fewer measurements to produce a signal with similar MSE. }
\label{speech_imputation}
\end{center}
\vskip -0.2in
\end{figure}

\begin{figure}[t]
\vskip 0.2in
\begin{center}
\centerline{\includegraphics[width=\columnwidth]{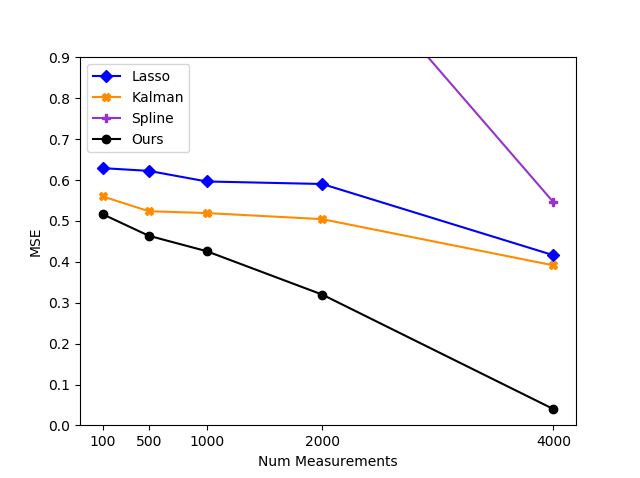}}
\caption{Test loss for imputation on an artificial chirp signal with a 500 hz frequency sweep with $n=16,384$ and varying numbers of measurement by our method, Lasso in the DCT basis, Kalman state-space imputation, and spline interpolation \cite{Moritz2017}. Our method greatly outperforms Kalman imputation and Lasso for all measurement levels.}
\label{chirp_imputation}
\end{center}
\vskip -0.2in
\end{figure}

\subsection{Data}
To test our methods accurately, we use three different types of univariate time series data with observations recorded at uniformly spaced intervals. First, we consider audio data, since they are naturally occurring time-varying signals. We use wav audio data subsampled to 8.192 kHz and clipped to two seconds long. Next, we consider sensor measurement data, which is a typical class of univariate time series used in recovery analysis. Finally, we generate chirp signals with linear frequency sweep to test performance on artificial series.

The data we use of each type are:

\begin{itemize}
    \item \textbf{Audio}: a recording of a humpback whale call \cite{knapp}, a drum beat, and a portion of a speech given by Bill Clinton.
    \item \textbf{Sensor}: recordings of NO2, O3, and CO levels in the air, taken hourly from an Italian town center between 04/09/2004 and 05/25/2004 \cite{DeVito2008, Dua2017}.
    \item \textbf{Chirp}: chirp signals with linear frequency sweep from 750 to 650 hz, 750 to 450 hz, and 750 to 250 hz.
\end{itemize}
The audio and chirp signals have length $n=16,384$ and the sensor data have length $n=1024$. We linearly normalize all signals to have range [-1,1] before processing.

\begin{figure*}[ht]
\vskip 0.2in
\begin{center}
\centerline{\includegraphics[width=\textwidth]{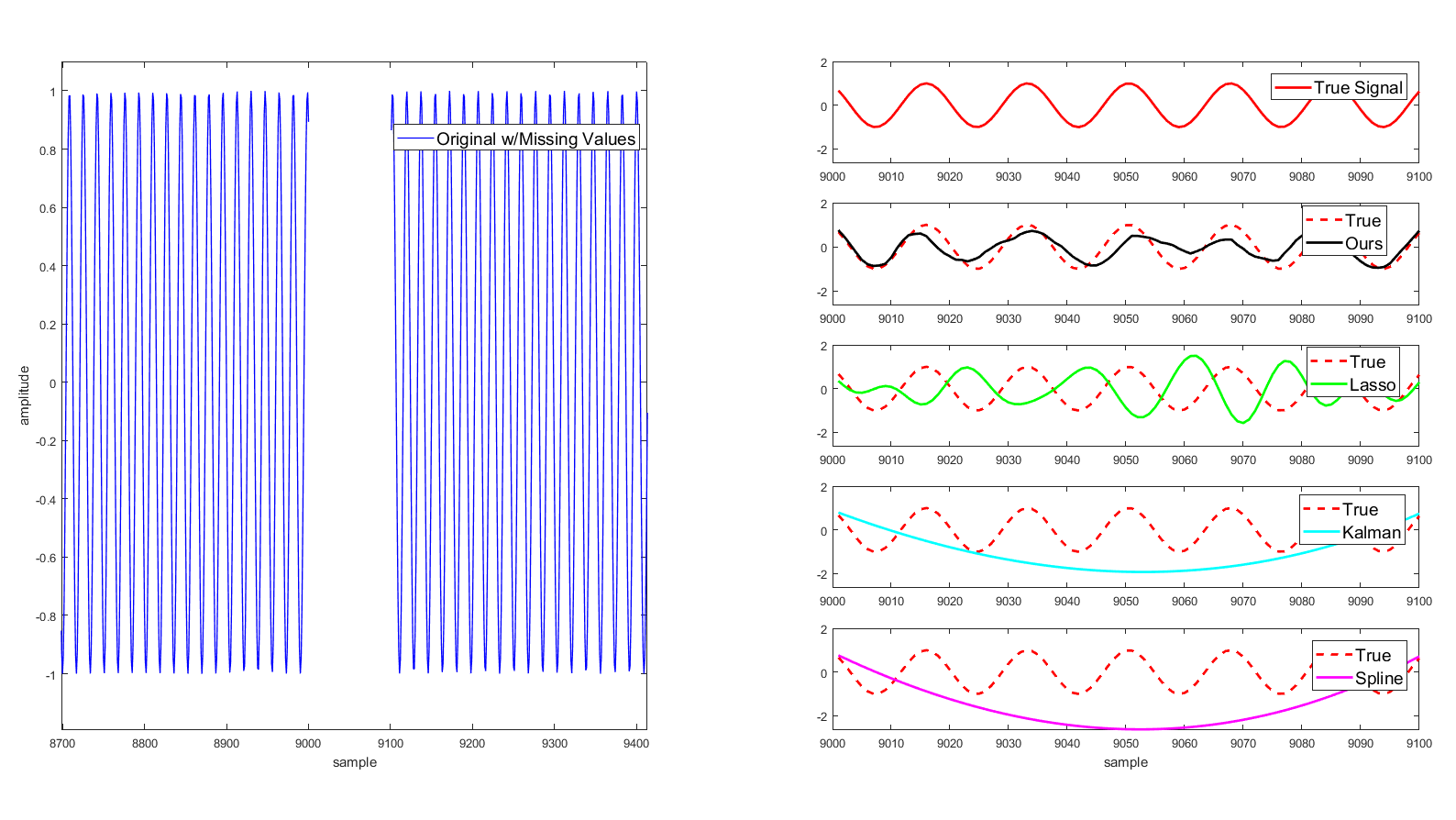}}
\caption{Imputation of a 100-sample section of a larger $n=16,384$ chirp signal by Lasso in the DCT basis, Kalman state-space imputation, spline interpolation, and our method. The original signal with missing values is on the left and the imputed values are on the right \cite{Moritz2017}. It is apparent from the reconstructions that only our method and Lasso properly capture the original shape of the signal, and only our reconstruction is in proper phase with the original signal. MSE values for each reconstruction are - ours: 0.483, Lasso: 1.787, Kalman: 14.084, Spline: 14.618.} 
\label{imputation_vis}
\end{center}
\vskip -0.2in
\end{figure*}

\subsection{Experimental Setup}
We treat four linear inverse problems on one-dimensional time series: imputation, recovery from random Gaussian projections, recovery from underdetermined discrete cosine transform measurements, and denoising. The first three problems deal with recovering a signal from underdetermined measurements with no noise (i.e. $m < n$ and $\eta[i]=0$ $\forall{i}$ s.t. $1\leq i \leq m$) and the last deals with the canonical problem of removing additive noise.

The first problem we test is imputation of missing time series observations. Imputation is one of the most commonly-studied forms of time series recovery since it has obvious applications for the real-world problem of missing data. The rows of measurement matrix $A$ are constructed by randomly sampling $m$ rows of the $n\times{n}$ identity matrix, where $m < n$.

Next we consider the case where $y$ is measured from underdetermined Gaussian projections of $x$. This is a common mode of compressed sensing used to evaluate signal reconstruction algorithms. The entries of $A$ are random iid Gaussian, scaled inversely by the square root of the number of observed measurements, $m$. 

Third, we test underdetermined DCT measurements, that is, the entries of $y$ are a random subset of the coefficients of the discrete cosine transform of the original signal $x$. This method tests how DIP performs at recovering a signal from projected measurements in a known domain. We construct $A$ by randomly sampling $m$ rows of the $n\times{n}$ DCT matrix, where $m < n$. 

Finally, we test our method's ability to remove additive white Gaussian noise (AWGN) from the original signal. The denoising of univariate time series is one of the most studied inverse signal problems and is a common application for recent model-based linear inverse methods \cite{Kohler2005}. In this case, $A$ is simply the $n\times{n}$ identity matrix and $\eta \in \mathbb{R}^{m}$ is a vector whose entries are iid random Gaussian with mean 0 and varying standard deviation. We test only the blind case, in which the denoising method does not know the level or nature of the noise added.

Our study implements three methods of regularization: total variation loss, $l_2$ weight decay, and early stopping. We implement total variation and weight decay as described in section~\ref{CNNsection} and hold these parameters constant for all tests. We implement early stopping by optimizing our network for 3000 steps for the imputation, compressed sensing, and DCT cases, and 300 steps for the denoising case to prevent overfitting to the noisy signal. We explore the effects of regularization more closely in the results section. 

\begin{table*}[t]
\caption{Blind denoising test loss results of our method, Lasso in the DCT basis, sym4 wavelet denoising, and Wiener filter on hourly air quality sensor time series from an Italian town \cite{DeVito2008}. Univariate time series of O3, NO2, and CO levels in the air were perturbed with AWGN with 0 mean and standard deviations 0.1,0.15, and 0.2.}
\label{denoising-table}
\vskip 0.2in
\begin{center}
\begin{small}
\begin{sc}
\begin{tabular}{@{}llllllllll@{}}
\toprule
Method        & \multicolumn{3}{c}{O3}                                                                                        & \multicolumn{3}{c}{NO2}                                                                                       & \multicolumn{3}{c}{CO}                                                                                        \\ \cmidrule(lr){2-4}\cmidrule(lr){5-7}\cmidrule(lr){8-10}
        & \multicolumn{1}{c}{$\sigma = 0.1$} & \multicolumn{1}{c}{$\sigma = 0.15$} & \multicolumn{1}{c}{$\sigma = 0.2$} & \multicolumn{1}{c}{$\sigma = 0.1$} & \multicolumn{1}{c}{$\sigma = 0.15$} & \multicolumn{1}{c}{$\sigma = 0.2$} & \multicolumn{1}{c}{$\sigma = 0.1$} & \multicolumn{1}{c}{$\sigma = 0.15$} & \multicolumn{1}{c}{$\sigma = 0.2$} \\ \midrule
Ours     & \textbf{0.0087}                    & \textbf{0.0161}                     & \textbf{0.0249}                    & \textbf{0.0091}                    & \textbf{0.0177}                     & \textbf{0.0220}                    & \textbf{0.0091}                    & \textbf{0.0170}                     &\textbf{0.0221}                    \\
Lasso   & 0.0574                             & 0.0673                              & 0.0852                             & 0.0570                             & 0.0695                              & 0.0817                             & 0.1151                             & 0.1276                              & 0.1412                             \\
Wavelet & 0.0114                             & 0.0188                              & 0.0283                             & 0.0109                             & 0.0201                              & 0.0283                             & 0.0109                             & 0.0183                              & 0.0279                             \\
Wiener  & 0.0602                             & 0.0659                              & 0.0747                             & 0.0598                             & 0.0665                              & 0.0716                             & 0.0791                             & 0.0838                              & 0.0912                             \\
\bottomrule
\end{tabular}
\end{sc}
\end{small}
\end{center}
\vskip -0.1in
\end{table*}

\subsection{Baseline Methods}
We compare our model to baseline algorithms for the four inverse problems we test by comparing the MSE of each algorithm's reconstruction vs. the original time series. For the imputation case specifically, we only compute the MSE between the imputed observations (i.e. the observations which each algorithm must predict) and their true values. We ignore the original observations from $x$ that were kept in the measurements $y$ after transformation by $A$.  We define this imputation loss as 

\begin{equation}
MSE_{imputation} = \frac{1}{|S|} \sum_{i \in S} (x{[i]} - \hat{x}{[i]})^2,
\end{equation}
\begin{center}
    $S =$ \{indices of missing observations\},
\end{center}

where $\hat{x}$ is the algorithm output, e.g. $G(z,w)$ for our method. For each test, we perform five random restarts of the network and average the five resultant MSE values to get a final value. 

We use several baselines to compare performance to our method:

\begin{itemize}
\item \textbf{Lasso}: we use scikit-learn's implementation of Lasso with a transform to the DCT basis as a baseline for all four inverse problems \cite{scikit-learn}. We set $\alpha$ = $10^{-5}$ as the best parameter from a search over several possible values. 
\item \textbf{D-VAMP}: we compare our performance in the DCT and Gaussian modes to D-VAMP \cite{Metzler2016, Rangan2017}. We use a Wiener filter with a window of 5 and a non-local means (NLM) filter as the denoisers. We choose the non-local means filter since it performs only slightly worse than BM3D-AMP for compressed sensing tasks, and a 1D implementation of BM3D is not currently available. Like our method, NLM-VAMP in these experiments extends a recovery method originally intended for images to univariate time series. 
\item \textbf{Kalman Imputation}: we use Kalman smoothing on a structural time series model, as implemented in imputeTS, as a baseline for imputation \cite{Moritz2017}.
\item \textbf{Spline Interpolation}:we use the imputeTS implementation of piecewise-polynomial spline interpolation as a baseline for imputation \cite{Moritz2017}.
\item \textbf{Wavelet Denoising}: we use the Matlab implementation of wavelet denoising with the default sym4 wavelet and Bayes thresholding.
\item \textbf{Wiener Filtering}:we use the Matlab implementation of Wiener filtering with a window size of five to compare denoising performance.
\end{itemize}

\section{Results}
In the case of imputation, Gaussian measurements, and DCT measurements, we perform trials with $m =$ 100, 500, 1000, 2000, and 4000 for the audio and chirp signals ($n = 16384$), and $m =$ 10, 25, 50, 75, and 150 for the air quality data ($n = 1024$). For denoising, we test AWGN with mean zero and standard deviation 0.1, 0.15, and 0.2. 

\subsection{Random Gaussian Measurements}
Our method outperformed all baselines for recovering natural audio signals with $n=16,384$ observations from random Gaussian projections with $m=100,500,1000,2000,$ and $4000$ measurements. Our method requires up to $4\times$ fewer measurements than Lasso in the DCT basis to reconstruct an audio time series of a drum beat, as seen in Figure~\ref{beat_cs}. NLM-VAMP slightly outperforms our method for higher numbers of measurements, but our method produces far more accurate reconstructions for fewer measurements. Our approach also outperforms all baselines in reconstructing artificial chirp signals for smaller numbers of measurements. 

On air quality data with $n=1024$ observations and $m=10,25,50,75,$ and $150$ measurements, our approach outperforms all baselines for 50 or fewer measurements. For higher numbers of measurements, Lasso in  the DCT basis outperforms our approach as observed in Figure~\ref{co_cs}.  

\subsection{DCT Measurements}
For DCT measurements of $m=100,500,1000,2000,$ and $4000$ on natural audio signals, our approach performed the best of all methods along with Lasso in DCT basis. As seen in Table~\ref{DCT-table}, NLM-VAMP did not converge for any of the audio signals and Wiener-Vamp performed well on only one time series. 

Our method and Lasso were the best methods on air quality data and artificial chirp signals as well. NLM-VAMP did not converge for any measurement level on any signal and Wiener-VAMP did not converge for many measurement ranges for air quality sensor data or chirp signals.

\begin{figure}[t]
\vskip 0.2in
\begin{center}
\centerline{\includegraphics[width=9.5cm]{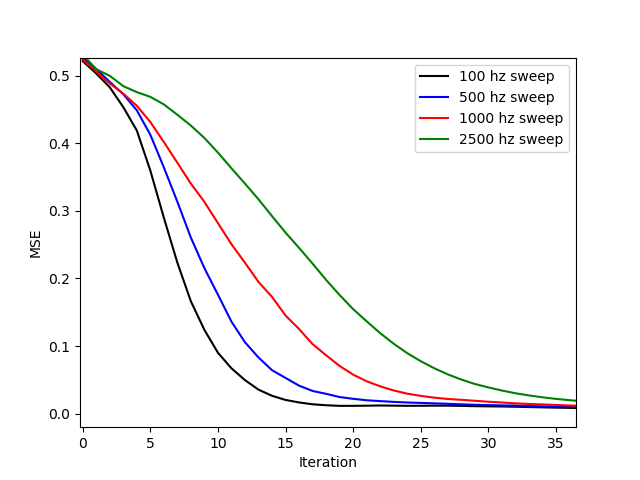}}
\caption{Training loss for chirp signals of varying complexity, ranging from a 100 hz sweep in frequency to a 2500 hz sweep. The network converges more slowly for time series with higher frequency components, which indicates that our network may tend toward local smoothness in its reconstructions. We may tune the total variation parameter to admit more non-smooth signals, but then likely admit more noise as well. On the other hand, we may tune the total variation parameter to produce signals that are even more smooth, but then dampen the network's ability to reproduce more complex signals.}
\label{signal_complexity}
\end{center}
\vskip -0.2in
\end{figure}

\subsection{Imputation}
Our method outperformed the Kalman state-space method at imputing audio signals, as shown in Figure~\ref{speech_imputation}, where our method requires up to $4\times$ fewer measurements than Kalman. Our approach also performed better at imputing missing observations in artificial chirp signals as seen Figure~\ref{chirp_imputation}. However, Kalman imputation required up to $2\times$ fewer measurements than our approach for air quality data. 

Figure~\ref{imputation_vis} shows how each method imputed a 100-sample gap in a larger $n=16,384$ chirp signal, with the original values shown in red and ours in black. Our method had the least MSE of all the reconstructions. Only our method and Lasso correctly reconstruct the shape of the signal, and only our reconstruction is properly in phase with the original signal.  
\subsection{Denoising}
To prevent overfitting to the noisy signal during blind denoising, we decreased the training iterations for our CNN to 300 and reduced the number of filters per layer to 16 for noise with $\sigma=0.15$ and 8 for $\sigma=0.2$.

Our method outperformed all baselines for denoising air quality time series as seen in Table~\ref{denoising-table}. Our approach had better blind denoising performance than the wavelet filter on air quality data and performed second to the wavelet filter for denoising audio signals.

\subsection{Effects of Regularization}
We had to carefully tune our $l_2$ weight decay, total variation, and early stopping iteration to prevent our network from overfitting to given measurements. We tuned our weight decay and total variation parameters and kept them constant for all of our experiments, but changed our stopping iteration for denoising. While DIP does have high impedance to reconstructing noise, it does fit a noisy signal given enough iterations, as shown in Figure~\ref{noise_fitting}, so stopping our network training earlier than the other experimental cases is necessary.

As we see in Figure~\ref{signal_complexity}, DIP fits less complex signals faster than more complex ones while training. While we could tune our total variation parameter to allow complex signals that are less piecewise smooth or to only allow very simple, smooth images, we choose to keep the total variation parameter constant to reduce the number of regularization variables to just one, the number of training iterations. 

\section{Conclusion}
We demonstrate the ability of deep image prior (DIP) methods to model and reconstruct 
one-dimensional natural signals from various types of linear measurement processes. 
Our method relies on the inductive bias of one-dimensional convolutional neural networks we design and also on optimized early stopping, total variation and weight decay regularization. 

We show results that are near-matching or outperforming state of the art methods for natural signals of various types and over four different types of measurement processes. 

 One-dimensional DIP requires up to $4\times$ fewer random Gaussian measurements than Lasso in the DCT basis to recover audio signals, performs as well as a Kalman state-space imputation model on a variety of data, and is more effective than sym4 wavelet filtering for blind denoising of air quality data. We further expand on the role of weight decay, total variation, and early stopping as forms of regularization.      

Future extension of our work includes extending it to multiple time signals, 
performing phase retrieval ~\cite{hand2018phase} or applying our temporal DIP methods to other problems~\cite{kothari2018random}, and combining sparse and deep priors for time-signals \cite{dhar2018modeling}.

\bibliography{Bibliography,references}
\bibliographystyle{icml2019}






\icmltitlerunning{DIP for Univariate Time Series Inverse Problems}


\onecolumn
\icmlkeywords{Compressed Sensing, Deep Image Prior, Time Series}

\vskip 0.3in

\section*{Appendix}

\begin{table}[ht]
\caption{Blind denoising MSE results of our DIP network, Lasso in the DCT basis, sym4 wavelet denoising, and Wiener Filter on audio data. Univariate time series of a whale call, speech, and a drum beat were perturbed with AWGN with 0 mean and standard deviations 0.1,0.15, and 0.2.}
\label{denoising-table}
\vskip 0.2in
\begin{center}
\begin{small}
\begin{sc}
\begin{tabular}{@{}lllllllllll@{}}
\toprule
Method        & \multicolumn{3}{c}{Whale}                                                                                        & \multicolumn{3}{c}{Speech}                                                                                       & \multicolumn{3}{c}{Beat}                                                                                        \\ \cmidrule(lr){2-4}\cmidrule(lr){5-7}\cmidrule(lr){8-10}
        & \multicolumn{1}{l}{$\sigma = 0.1$} & \multicolumn{1}{l}{$\sigma = 0.15$} & \multicolumn{1}{l}{$\sigma = 0.2$} & \multicolumn{1}{l}{$\sigma = 0.1$} & \multicolumn{1}{l}{$\sigma = 0.15$} & \multicolumn{1}{l}{$\sigma = 0.2$} & \multicolumn{1}{l}{$\sigma = 0.1$} & \multicolumn{1}{l}{$\sigma = 0.15$} & \multicolumn{1}{l}{$\sigma = 0.2$} \\ \midrule
Ours     & 0.0038          & 0.0058          & 0.0133          & 0.0059          & 0.0094          & 0.0284          & 0.0029          & 0.0038          & 0.0043          \\

Lasso   &0.0499          & 0.0522          & 0.0584          & 0.0104          & 0.0131          & 0.0197          & 0.0062          & 0.0088          & 0.0151          \\

Wavelet & \textbf{0.0032} & \textbf{0.0054} & \textbf{0.0075} & \textbf{0.0047} & \textbf{0.0078} & \textbf{0.0116} & \textbf{0.0008} & \textbf{0.0012} & \textbf{0.0017} \\

Wiener  & 0.0260          & 0.0297          & 0.0362          & 0.0138          & 0.0184          & 0.0251          & 0.0052          & 0.0096          & 0.0171         \\
\bottomrule
\end{tabular}
\end{sc}
\end{small}
\end{center}
\vskip 0.2in
\end{table}

\begin{table}[ht]
\caption{Blind denoising MSE results of our DIP network, Lasso in the DCT basis, sym4 wavelet denoising, and Wiener Filter on artificial chirp signals. Univariate time series of chirps with a 500 hz, 300 hz, and 100 hz frequency shift were perturbed with AWGN with 0 mean and standard deviations 0.1,0.15, and 0.2.}
\label{denoising-table}
\vskip 0.2in
\begin{center}
\begin{small}
\begin{sc}
\begin{tabular}{@{}lllllllllll@{}}
\toprule
Method        & \multicolumn{3}{c}{500 hz}                                                                                        & \multicolumn{3}{c}{300 hz}                                                                                       & \multicolumn{3}{c}{100 hz}                                                                                        \\ \cmidrule(lr){2-4}\cmidrule(lr){5-7}\cmidrule(lr){8-10}
        & \multicolumn{1}{l}{$\sigma = 0.1$} & \multicolumn{1}{l}{$\sigma = 0.15$} & \multicolumn{1}{l}{$\sigma = 0.2$} & \multicolumn{1}{l}{$\sigma = 0.1$} & \multicolumn{1}{l}{$\sigma = 0.15$} & \multicolumn{1}{l}{$\sigma = 0.2$} & \multicolumn{1}{l}{$\sigma = 0.1$} & \multicolumn{1}{l}{$\sigma = 0.15$} & \multicolumn{1}{l}{$\sigma = 0.2$} \\ \midrule
Ours     & 0.0092          & 0.0109          & 0.0689          & 0.0086          & 0.0087          & 0.0290          & 0.0078          & 0.0071          & 0.0114          \\

Lasso   &0.0049          & \textbf{0.0086} & \textbf{0.0161} & \textbf{0.0032} & \textbf{0.0062} & \textbf{0.0136} & \textbf{0.0015} & \textbf{0.0040} & \textbf{0.0106}          \\

Wavelet & \textbf{0.0045} & 0.0104          & 0.0189          & 0.0053          & 0.0108          & 0.0215          & 0.0060          & 0.0128          & 0.0213           \\

Wiener  & 0.1693          & 0.1793          & 0.1893          & 0.1703          & 0.1788          & 0.1894          & 0.1696          & 0.1789          & 0.1882         \\
\bottomrule
\end{tabular}
\end{sc}
\end{small}
\end{center}
\vskip -0.1in
\end{table}

\begin{table*}[t]
\caption{MSE results for reconstructing a signal from $m=10,25,50,75,$ and $150$  DCT coefficient measurements of air sensor time series with $n = 1024$. We compare our DIP method to Lasso in the DCT basis, NLM-VAMP, and Wiener-VAMP. The best MSE value for each test is bolded. DNC indicates that the algorithm did not converge.}
\label{DCT-table}
\vskip 0.2in
\begin{center}
\begin{small}
\begin{sc}
\begin{tabular}{@{}llllll@{}}
\toprule
Method      & \multicolumn{5}{c}{O3}                                                                                                                                                                         \\\cmidrule(lr){2-6}
            & \multicolumn{1}{c}{$m =10$}         & \multicolumn{1}{c}{$m =25$}         & \multicolumn{1}{c}{$m =50$}        & \multicolumn{1}{c}{$m =75$}        & \multicolumn{1}{c}{$m =150$}        \\\midrule
Ours         &0.2185          & \textbf{0.1835} & 0.1594          & 0.1519          & 0.1433          \\
Lasso       &\textbf{0.2063} & 0.2036 & \textbf{0.1578} & \textbf{0.1496} & \textbf{0.1396} \\
NLM-VAMP    &DNC             & DNC    & DNC             & DNC             & DNC             \\
Wiener-VAMP &DNC             & DNC    & DNC             & DNC             & 0.1485                   \\\midrule
            & \multicolumn{5}{c}{NO2}                                                                                                                                                                        \\\cmidrule(lr){2-6}
            & $m =10$                             & $m =25$                             & $m=50$                            & $m =75$                            & $m =150$                            \\\midrule
Ours         & \textbf{0.2060} & \textbf{0.2051} & 0.1594          & 0.1533          & 0.1372          \\
Lasso       & 0.2062          & 0.2059          & \textbf{0.1552} & \textbf{0.1492} & \textbf{0.1297} \\
NLM-VAMP    &DNC             & DNC             & DNC             & DNC             & DNC             \\
Wiener-VAMP & DNC             & DNC             & DNC             & DNC             & 0.1400 \\\midrule
            & \multicolumn{5}{c}{CO}                                                                                                                                                                       \\\cmidrule(lr){2-6}
            & $m =10$                             & $m =25$                             & $m =50$                            & $m =75$            & $m=150$                            \\\midrule
Ours         & \textbf{0.2496} & \textbf{0.2474} & \textbf{0.2468} & \textbf{0.2307} & 0.1161          \\
Lasso       & 0.2540          & 0.2534          & 0.2522          & 0.2471          & \textbf{0.1107} \\
NLM-VAMP    & DNC             & DNC             & DNC             & DNC             & DNC             \\
Wiener-VAMP & DNC             & 0.3126          & 0.2747          & 0.2695          & DNC            \\                             
\bottomrule
\end{tabular}
\end{sc}
\end{small}
\end{center}
\vskip -0.1in
\end{table*}

\begin{table*}[t]
\caption{MSE results for reconstructing a signal from $m=100,500,1000,2000,$ and $4000$  DCT coefficient measurements of artificial chirp signals with $n = 16,384$ and linear phase shifts of 500 hz, 300 hz, and 100 hz. We compare our DIP method to Lasso in the DCT basis, NLM-VAMP, and Wiener-VAMP. The best MSE value for each test is bolded. DNC indicates that the algorithm did not converge.}
\label{DCT-table}
\vskip 0.2in
\begin{center}
\begin{small}
\begin{sc}
\begin{tabular}{@{}llllll@{}}
\toprule
Method      & \multicolumn{5}{c}{500 hz}                                                                                                                                                                         \\\cmidrule(lr){2-6}
            & \multicolumn{1}{c}{$m =100$}         & \multicolumn{1}{c}{$m =500$}         & \multicolumn{1}{c}{$m =1000$}        & \multicolumn{1}{c}{$m =2000$}        & \multicolumn{1}{c}{$m =4000$}        \\\midrule
Ours         & 0.5013          & 0.5003          & 0.4844          & 0.4552          & 0.3919          \\
Lasso       &\textbf{0.4973} & \textbf{0.4872} & \textbf{0.4676} & \textbf{0.4367} & \textbf{0.3725} \\
NLM-VAMP    & DNC                                & DNC                                  & DNC                                  & DNC                                  & DNC                                   \\
Wiener-VAMP & DNC             & DNC             & 0.6357          & 0.4625          & 0.4055           \\\midrule
            & \multicolumn{5}{c}{300 hz}                                                                                                                                                                        \\\cmidrule(lr){2-6}
            & $m =100$                             & $m =500$                             & $m =1000$                            & $m =2000$                            & $m =4000$                            \\\midrule
Ours         & 0.4998          & 0.4949          & 0.4749          & 0.4506          & 0.4053          \\
Lasso       & \textbf{0.4972} & \textbf{0.4896} & \textbf{0.4657} & \textbf{0.4356} & \textbf{0.3812} \\
NLM-VAMP    & DNC                                  & DNC                                  & DNC                                  & DNC                                  & DNC                                  \\
Wiener-VAMP & DNC             & DNC             & 0.6437          & 0.4619          & 0.4177                                \\\midrule
            & \multicolumn{5}{c}{100 hz}                                                                                                                                                                       \\\cmidrule(lr){2-6}
            & $m =100$                             & $m =500$                             & $m =1000$                            & $m =2000$                            & $m =4000$                            \\\midrule
Ours         & 0.5009          & \textbf{0.4858} & \textbf{0.4517} & \textbf{0.4255} & 0.3923          \\
Lasso       & \textbf{0.4959} & 0.4864          & 0.4551          & 0.4261          & \textbf{0.3665} \\
NLM-VAMP    & DNC                                  & DNC                                  & DNC                                  & DNC                                  & DNC                                  \\
Wiener-VAMP & DNC             & DNC             & DNC             & 0.4562          & 0.3997      \\                             
\bottomrule
\end{tabular}
\end{sc}
\end{small}
\end{center}
\vskip -0.1in
\end{table*}

\begin{figure*}[th]
\vskip 0.2in
\begin{center}
\centerline{\includegraphics[width=10cm]{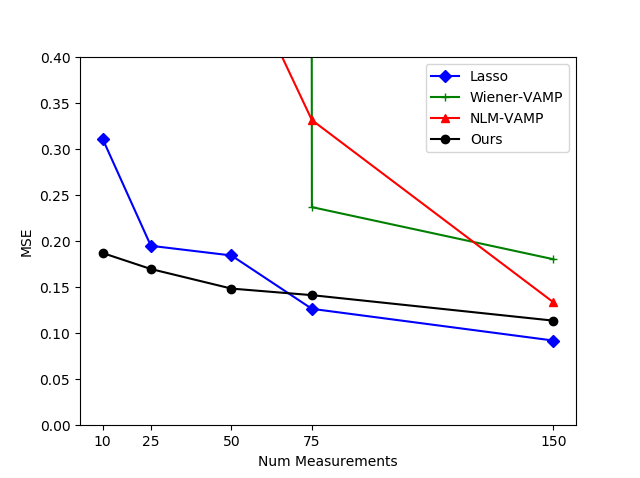}}
\caption{Test loss for recovery from random Gaussian  projections on hourly sensor readings of NO2 in the air with $n=1024$ and varying numbers of measurement by our method, Lasso in the DCT basis, NLM-VAMP, and Wiener-VAMP.}
\label{co_cs}
\end{center}
\vskip -0.2in
\end{figure*}

\begin{figure*}[th]
\vskip 0.2in
\begin{center}
\centerline{\includegraphics[width=10cm]{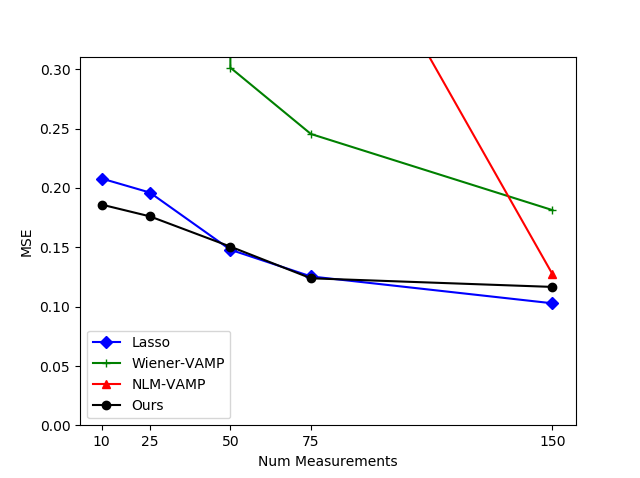}}
\caption{Test loss for recovery from random Gaussian  projections on hourly sensor readings of O3 in the air with $n=1024$ and varying numbers of measurement by our method, Lasso in the DCT basis, NLM-VAMP, and Wiener-VAMP.}
\label{co_cs}
\end{center}
\vskip -0.2in
\end{figure*}

\begin{figure*}[th]
\vskip 0.2in
\begin{center}
\centerline{\includegraphics[width=10cm]{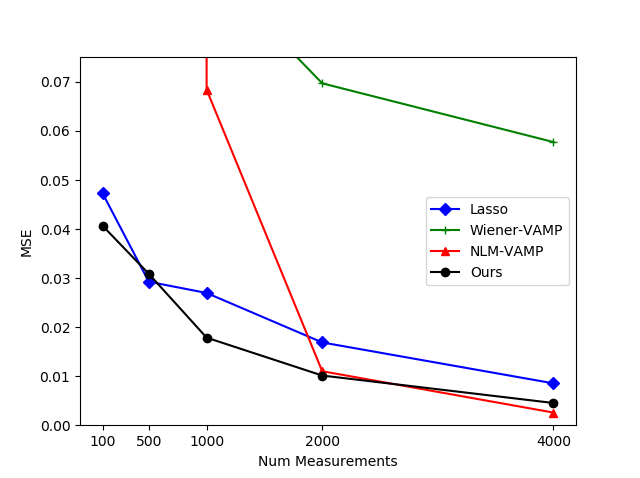}}
\caption{Test loss for recovery from random Gaussian  projections on an audio signal of a whale call with $n=16,384$ and varying numbers of measurement by our method, Lasso in the DCT basis, NLM-VAMP, and Wiener-VAMP.}
\label{beat_cs}
\end{center}
\vskip -0.2in
\end{figure*}

\begin{figure*}[th]
\vskip 0.2in
\begin{center}
\centerline{\includegraphics[width=10cm]{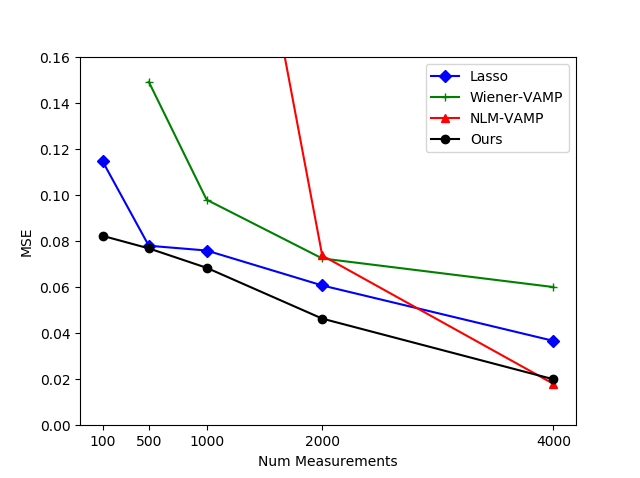}}
\caption{Test loss for recovery from random Gaussian  projections on an audio signal of speech with $n=16,384$ and varying numbers of measurement by our method, Lasso in the DCT basis, NLM-VAMP, and Wiener-VAMP.}
\label{beat_cs}
\end{center}
\vskip -0.2in
\end{figure*}

\begin{figure*}[th]
\vskip 0.2in
\begin{center}
\centerline{\includegraphics[width=10cm]{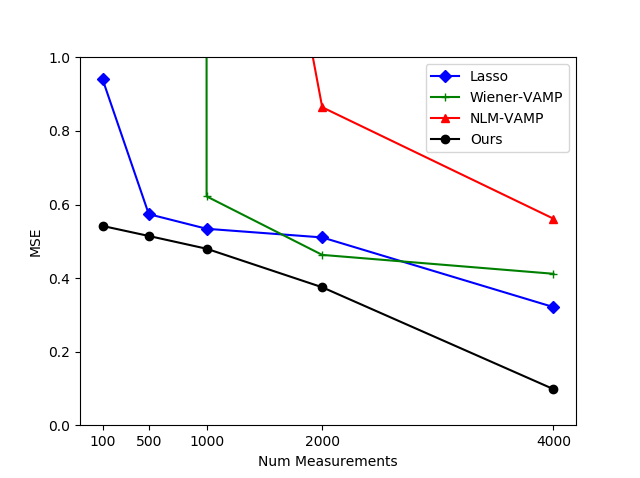}}
\caption{Test loss for recovery from random Gaussian  projections on an artificial chirp signal with a 500 hz frequency sweep with $n=16,384$ and varying numbers of measurement by our method, Lasso in the DCT basis, NLM-VAMP, and Wiener-VAMP.}
\label{beat_cs}
\end{center}
\vskip -0.2in
\end{figure*}

\begin{figure*}[th]
\vskip 0.2in
\begin{center}
\centerline{\includegraphics[width=10cm]{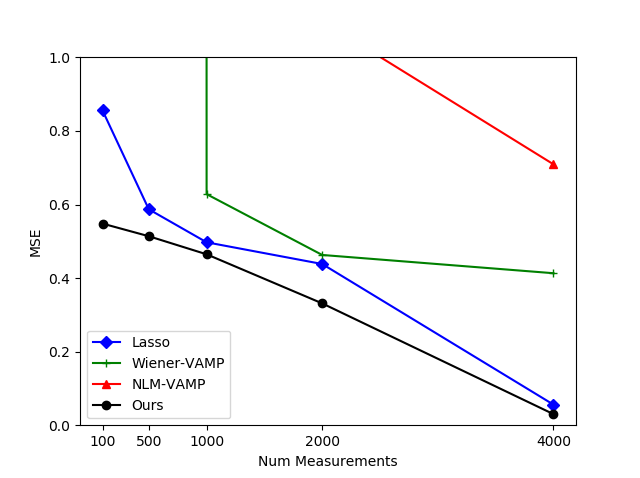}}
\caption{Test loss for recovery from random Gaussian  projections on an artificial chirp signal with a 300 hz frequency sweep with $n=16,384$ and varying numbers of measurement by our method, Lasso in the DCT basis, NLM-VAMP, and Wiener-VAMP.}
\label{beat_cs}
\end{center}
\vskip -0.2in
\end{figure*}

\begin{figure*}[t]
\vskip 0.2in
\begin{center}
\centerline{\includegraphics[width=10cm]{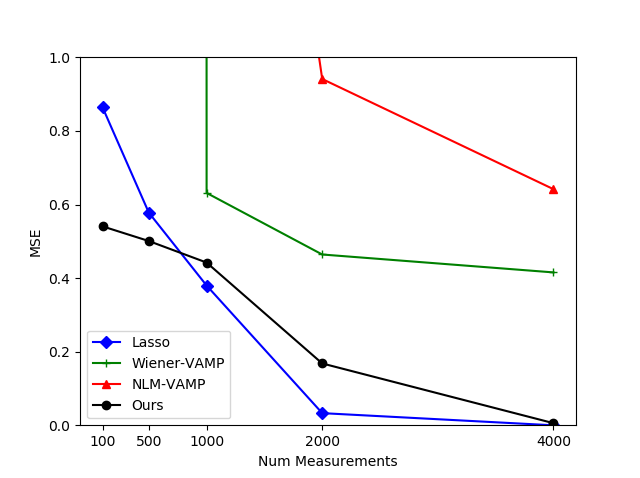}}
\caption{Test loss for recovery from random Gaussian  projections on an artificial chirp signal with a 100 hz frequency sweep with $n=16,384$ and varying numbers of measurement by our method, Lasso in the DCT basis, NLM-VAMP, and Wiener-VAMP.}
\label{beat_cs}
\end{center}
\vskip 2.5in
\end{figure*}


\begin{figure}[t]
\vskip 0.2in
\begin{center}
\centerline{\includegraphics[width=10cm]{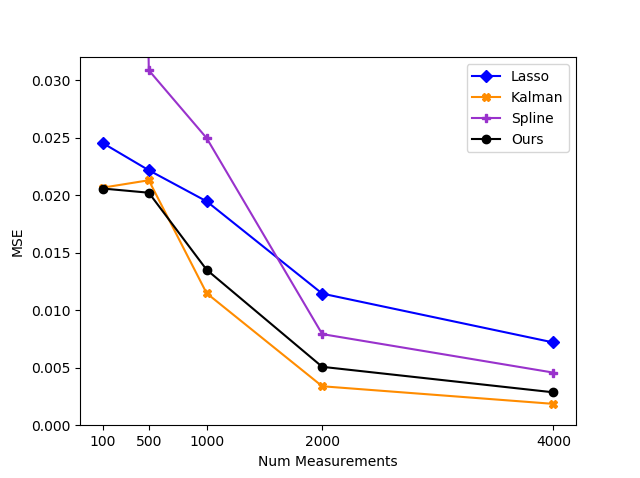}}
\caption{Test loss for imputation on an audio signal of a drum beat with $n=16,384$ and varying numbers of measurement by our method, Lasso in the DCT basis, Kalman state-space imputation, and spline interpolation.}
\label{speech_imputation}
\end{center}
\vskip -0.2in
\end{figure}

\begin{figure}[b]
\vskip 0.2in
\begin{center}
\centerline{\includegraphics[width=10cm]{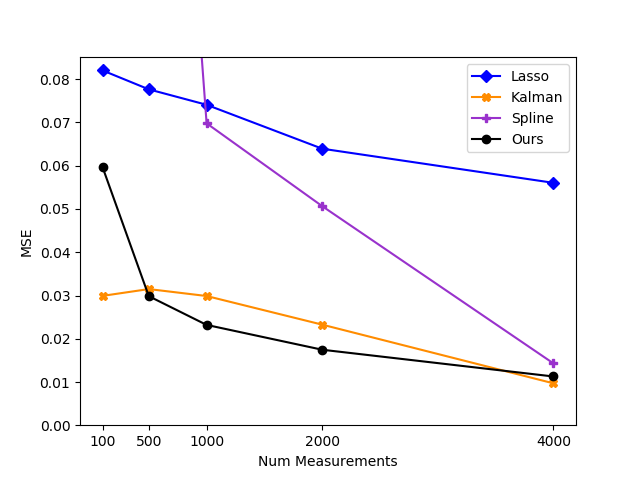}}
\caption{Test loss for imputation on an audio signal of a whale call with $n=16,384$ and varying numbers of measurement by our method, Lasso in the DCT basis, Kalman state-space imputation, and spline interpolation.}
\label{speech_imputation}
\end{center}
\vskip -0.2in
\end{figure}

\begin{figure}[t]
\vskip 0.2in
\begin{center}
\centerline{\includegraphics[width=10cm]{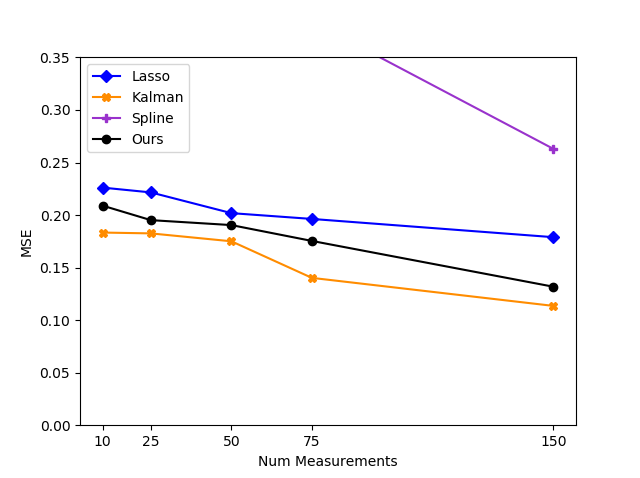}}
\caption{Test loss for imputation on hourly sensor readings of O3 in the air with $n=1024$ and varying numbers of measurement by our method, Lasso in the DCT basis, Kalman state-space imputation, and spline interpolation.}
\label{speech_imputation}
\end{center}
\vskip -0.2in
\end{figure}

\begin{figure}[b]
\vskip 0.2in
\begin{center}
\centerline{\includegraphics[width=10cm]{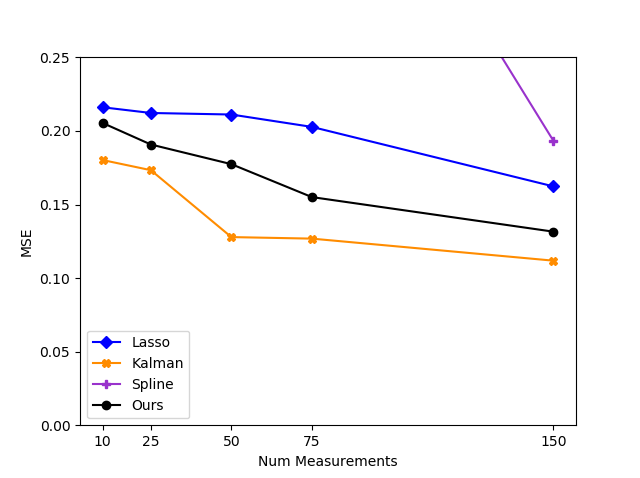}}
\caption{Test loss for imputation on hourly sensor readings of NO2 in the air with $n=1024$ and varying numbers of measurement by our method, Lasso in the DCT basis, Kalman state-space imputation, and spline interpolation.}
\label{speech_imputation}
\end{center}
\vskip -0.2in
\end{figure}

\begin{figure}[t]
\vskip 0.2in
\begin{center}
\centerline{\includegraphics[width=10cm]{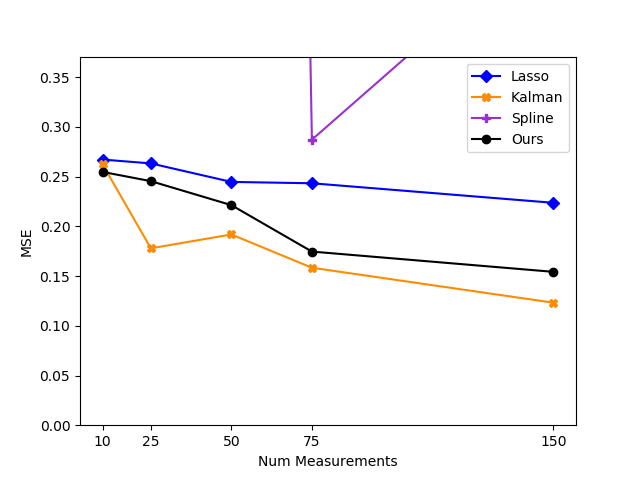}}
\caption{Test loss for imputation on hourly sensor readings of CO in the air with $n=1024$ and varying numbers of measurement by our method, Lasso in the DCT basis, Kalman state-space imputation, and spline interpolation.}
\label{speech_imputation}
\end{center}
\vskip -0.2in
\end{figure}

\begin{figure}[b]
\vskip 0.2in
\begin{center}
\centerline{\includegraphics[width=10cm]{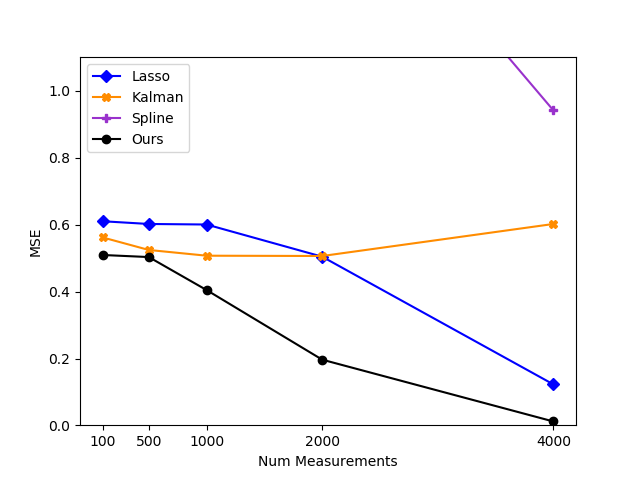}}
\caption{Test loss for imputation on an artificial chirp signal with a 300 hz frequency sweep with $n=16,384$ and varying numbers of measurement by our method, Lasso in the DCT basis, Kalman state-space imputation, and spline interpolation.}
\label{chirp_imputation}
\end{center}
\vskip -0.2in
\end{figure}

\begin{figure*}[t]
\vskip 0.2in
\begin{center}
\centerline{\includegraphics[width=10cm]{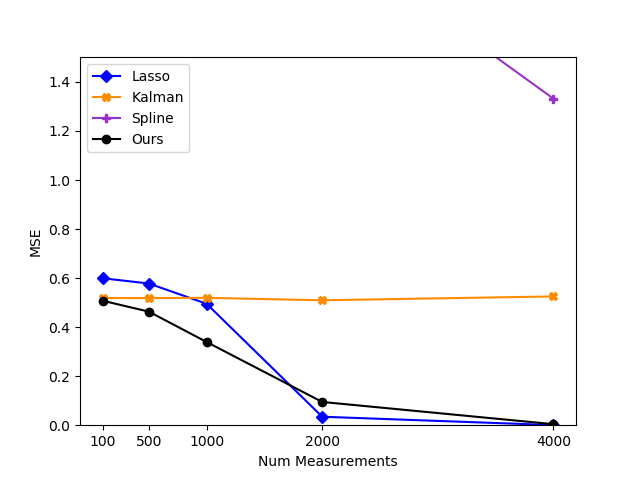}}
\caption{Test loss for imputation on an artificial chirp signal with a 100 hz frequency sweep with $n=16,384$ and varying numbers of measurement by our method, Lasso in the DCT basis, Kalman state-space imputation, and spline interpolation.}
\label{chirp_imputation}
\end{center}
\vskip -0.2in
\end{figure*}

\end{document}